\documentclass{article}
\pdfoutput=1

\PassOptionsToPackage{numbers, compress}{natbib}

\usepackage{notations}
 \usepackage[preprint]{neurips_2020}

\usepackage[utf8]{inputenc} 
\usepackage[T1]{fontenc}    
\usepackage[pagebackref=true,breaklinks=true,letterpaper=true,colorlinks,bookmarks=false,citecolor=blue,linkcolor=red]{hyperref}       
\usepackage{url}            
\usepackage{booktabs}       
\usepackage{amsfonts}       
\usepackage{nicefrac}       
\usepackage{microtype}      

\usepackage[pdftex]{graphicx}          
\usepackage{comment}
\usepackage{amsmath,amssymb} 
\usepackage{color}
\usepackage{colortbl}
\usepackage{geometry}
\usepackage[english]{babel}
\usepackage{algorithm}
\usepackage{adjustbox}
\usepackage{import}
\usepackage[noend]{algpseudocode}
\usepackage{relsize}
\usepackage{subcaption}
\usepackage{bm}
\usepackage{mathtools}
\usepackage{multirow}
\usepackage{array}
\usepackage{enumitem}
\usepackage[low-sup]{subdepth}
\usepackage{xspace}
\usepackage{bm}
\usepackage[font=small,labelfont=bf]{caption}
\usepackage{varwidth}
\newsavebox\tmpbox

\newcommand{\ours}{SKD\xspace}

\definecolor{mygray}{gray}{.9}
\title{Self-supervised Knowledge Distillation for \\ Few-shot Learning}

 \author{
    Jathushan Rajasegaran$^{1}$ \And Salman Khan$^{1}$ \And  Munawar Hayat$^{1}$ \AND Fahad Shahbaz Khan$^{1,2}$ \And  Mubarak Shah$^{3}$ \AND \tabularnewline
 \small $^{1}$Inception Institute of Artificial Intelligence, UAE \ \ \ \ \ $^{2}$CVL, Linköping University, Sweden \tabularnewline $^{3}$University of Central Florida, USA \tabularnewline
\texttt{\small \{first.lastname\}@inceptioniai.org}, \texttt{\small mshah@ucf.edu}  \tabularnewline
}

\begin{document}

\maketitle

\begin{abstract}

Real-world contains an overwhelmingly large number of object classes, learning all of which at once is infeasible. 
Few shot learning is a promising learning paradigm due to its ability to learn out of order distributions quickly with only a few samples.
Recent works~\cite{Dhillon2020A,tian2020rethink} show that simply learning a good feature embedding can outperform more sophisticated meta-learning and metric learning algorithms for few-shot learning. In this paper, we propose a simple approach to improve the representation capacity of deep neural networks for few-shot learning tasks. We follow a two-stage learning process: \emph{First}, we train a neural network to maximize the entropy of the feature embedding, thus creating an optimal output manifold using a self-supervised auxiliary loss. In the \emph{second} stage, we minimize the entropy on feature embedding by bringing self-supervised twins together, while constraining the manifold with student-teacher distillation. Our experiments show that, even in the first stage,  self-supervision can outperform current state-of-the-art methods, with further gains achieved by our second stage distillation process. Our codes are available at: \small{\url{https://github.com/brjathu/SKD}}\,.


\end{abstract}
\section{Introduction}
\label{sec:introduction}
Modern deep learning algorithms generally require a large amount of annotated data which is often laborious and expensive to acquire \cite{bengio2017deep,khan2018guide}. Inspired by the fact that humans can learn from only a few-examples, few-shot learning (FSL) offers a promising machine learning paradigm. FSL aims to develop models that can generalize to new concepts using only a few annotated samples (typically ranging from 1-5). Due to data scarcity and limited supervision, FSL remains a challenging problem.

Existing works mainly approach FSL using meta-learning \cite{finn2017model, li2017meta, jamal2019task, rusu2018metalearning, bertinetto2018meta, lee2019meta,ravichandran2019few} to adapt the base learner for the new tasks, or by enforcing margin maximizing constraints through metric learning \cite{koch2015siamese, sung2018learning, vinyals2016matching, snell2017prototypical}. In doing so, these FSL methods ignore the importance of intra-class diversity while seeking to achieve inter-class discriminability. In this work, instead of learning representations which are invariant to within class changes, we argue for an equivariant representation. Our main intuition is that major transformations in the input domain are desired to be reflected in their corresponding outputs to ensure output space diversity. By faithfully reflecting these changes in an equivariant manner, we seek to learn the true natural manifold of an object class.

We propose a two-stage self-supervised knowledge distillation (SKD) approach for FSL. Despite the availability of only few-shot labeled examples, we show that auxiliary self-supervised learning (SSL) signals can be mined from the limited data, and effectively leveraged to learn the true output-space manifold of each class. For this purpose, we take a direction in contrast to previous works which learn an invariant representation that maps augmented inputs to the same prediction. With the goal to enhance generalizability of the model, we {first} learn a \emph{Generation zero} (Gen-0) model whose output predictions are equivariant to the input transformations, thereby avoiding overfitting and ensuring heterogeneity in the prediction space. For example, when learning to classify objects in the first stage of learning, the self-supervision based learning objective ensures that the output logits are rich enough to encode the amount of rotation applied to the input images.

Once the generation zero network is trained to estimate the optimal output manifold, we perform knowledge distillation by treating the learned model as a teacher network and training a student model with the teacher's outputs. Different to first stage, we now enforce that the augmented samples and original inputs result in similar predictions to enhance between-class discrimination. The knowledge distillation mechanism therefore guides the \emph{Generation one} (Gen-1) model to develop two intuitive properties. \emph{First,} the output class manifold is diverse enough to preserve major transformations in the input, thereby avoiding overfitting and improving generalization. \emph{Second,} the learned relationships in the output space encode natural connections between classes e.g., two similar classes should have correlated predictions as opposed to totally independent categories considered in one-hot encoded ground-truths. Thus, by faithfully representing the output space via encoding inter-class relationships and preserving intra-class diversity, our approach learns improved representations for FSL.

\begin{figure}[t]
    \centering
    \begin{minipage}{0.6\textwidth}\vspace{-1em}
    \includegraphics[width=1\textwidth]{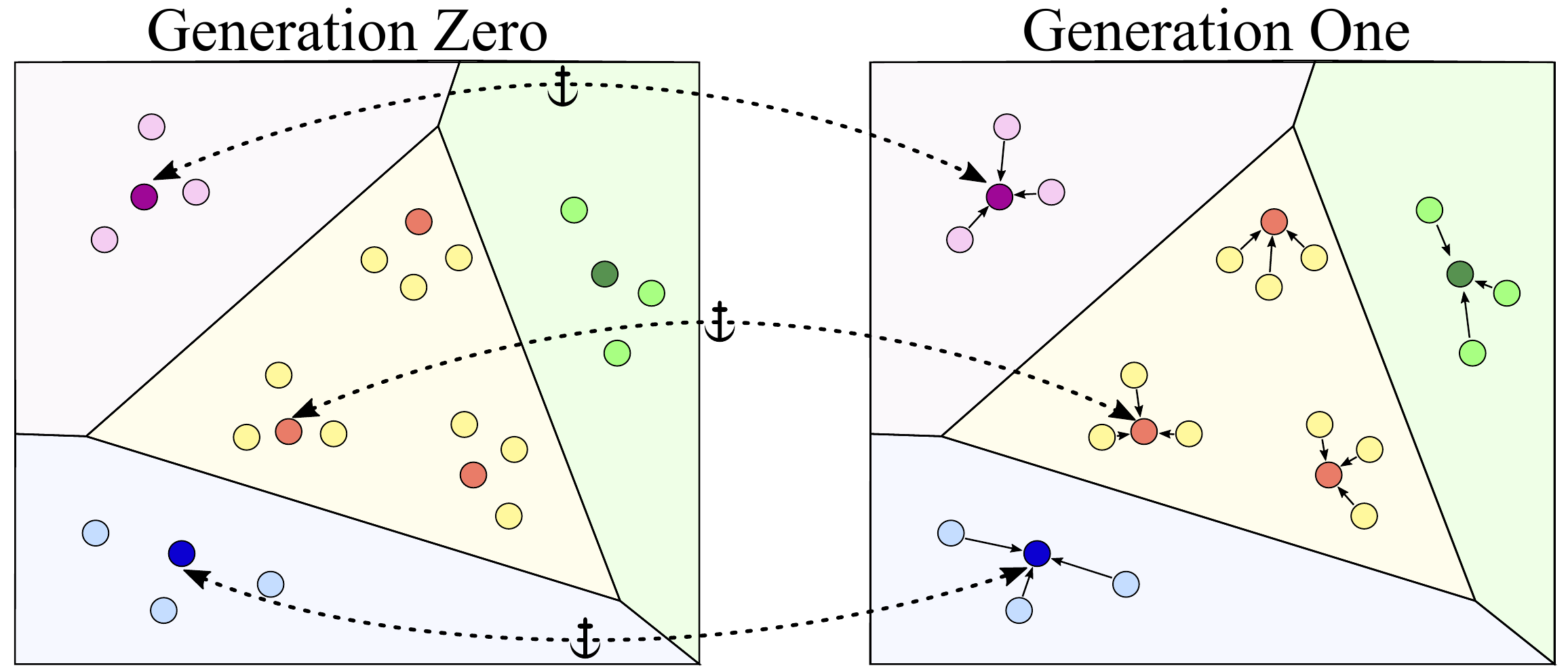}
    \end{minipage}\hfill
    \begin{minipage}{0.39\textwidth}
        \captionof{figure}{\emph{Self-supervised Knowledge Distillation} operates in two phases. In Gen-0, self-supervision is used to estimate the true prediction manifold, equivariant to input transformations. Specifically, we enforce the model to predict the amount of input rotation using only the output logits. In Gen-1, we force the original sample outputs to be the same as in Gen-0 (\emph{dotted lines}), while reducing its distance with its augmented versions to enhance discriminability.}
        \label{fig:main}
    \end{minipage}
\end{figure}

The following are the main contributions of this work (see Fig.~\ref{fig:main} for an overview):
\begin{itemize}[leftmargin=*,topsep=0pt,itemsep=0pt]
    \item Different to existing works that use SSL as an auxiliary task, we show the benefit of SSL towards enforcing diversity constraints in the prediction space with a simple architectural modification.  
    \item A dual-stage training regime which first estimates the optimal output manifold, and then minimizes the original-augmented pair distance while  anchoring the original samples to the learned manifold using a distillation loss.
    \item Extensive evaluations on four popular benchmark datasets with significant improvements on the FSL task.
\end{itemize}

\section{Related work}
\label{sec:related_work}

\textbf{Self-supervised learning (SSL):} This form of learning defines auxiliary learning tasks that can enhance model's learning capability without requiring any additional annotation effort. Generally, these surrogate tasks require a higher-level understanding, thereby forcing the learning agent to learn useful representations while solving the auxiliary tasks. The main difference in existing SSL techniques is regarding the way supervisory signal is obtained from the data. For example, \cite{gidaris2018unsupervised} learns useful representations by predicting the amount of rotation applied to an input image. Doersch~\emph{et al.} \cite{doersch2015unsupervised} train a CNN to predict the relative position of a pair of randomly sampled image patches. This idea is further extended to predict permutations of multiple image patches in \cite{noroozi2016unsupervised}.  Alternatively, image colorization and object counting were employed as pretext tasks to improve representation learning \cite{zhang2016colorful,noroozi2017representation}. Zhai~\emph{et al.} \cite{zhai2019s4l} propose an SSL approach in a semi-supervised setting where some labelled and many unlabelled examples were available. Different from these works, our approach uses self-supervision to enforce additional constraints in the classification space. Close to our work is a set of approaches that seek to learn representations that are invariant to image transformations and augmentations \cite{dosovitskiy2014discriminative, caron2018deep, chen2020simple}. In contrast, our approach does the exact opposite: we seek to learn an equivariant representation, so that the true natural manifold of an object class can be learned with only a few-examples.

\textbf{Few-shot learning (FSL):} There has been several efforts on FSL ranging from metric learning to meta learning methods. Metric learning methods commonly learn a metric space, in which the support set can be easily matched with the query set. For example, Koch~\emph{et al.}~\cite{koch2015siamese} use a Siamese network to learn a similarity metric to classify unknown classes, with the aid of a support set. Sung~\emph{et al.}~\cite{sung2018learning} use a relation module to learn the relationships between support set and the query image. Matching networks~\cite{vinyals2016matching} employ attention and memory to learn a network that matches support set to the query image. In addition,~\cite{snell2017prototypical} assigns the mean embedding as a prototype and minimizes the distance from it with rest of the samples in the query set. In contrast, we only use augmented pairs of an image to move their embeddings closer, while preserving their respective distances in the output space.

Another category of methods employ meta-learning to leverage from the knowledge acquired from the past tasks to learn new tasks. Finn~\emph{et al.}~\cite{finn2017model} proposed a popular model-agnostic meta-learning (MAML) framework, which finds better initialization weights that can be quickly adopted to a given support set. Building on \cite{finn2017model},  ~\cite{li2017meta, Flennerhag2020} use meta-learned preconditioning to redirect the gradient-flow to achieve better convergence. 
In addition to these works, LEO (Latent Embedding Optimization)~\cite{rusu2018metalearning} transforms network weights to a lower dimensional latent embedding space and applies MAML algorithm to scale to larger networks.
MetaOptNet~\cite{lee2019meta} employs an SVM to model meta-learning as a convex optimization problem which is solved using quadratic programming. 

Some recent works attribute the success of meta-learning to its strong feature representation capability rather than meta-learning itself \cite{raghu2019rapid}. Others~\cite{Dhillon2020A, tian2020rethink} show the effectiveness of a simple baseline by learning a strong embedding. This work is an effort along the same direction, and proposes a novel self-supervised knowledge distillation approach which can learn effective feature representations for FSL. 
The closest to our work is Gidaris~\emph{et al.}~\cite{gidaris2019boosting}, who use self-supervision to boost few-shot classification. However, \cite{gidaris2019boosting} simply uses self-supervision as an auxiliary loss for single training, while we use it to shape and constrain the learning manifold.
Architecture wise, we use a sequential self-supervision layer, while~\cite{gidaris2019boosting} has a parallel design. While \cite{gidaris2019boosting} does not have multiple generations, we further improve the representations in the second generation, by constraining the embedding space using distillation and bringing embeddings of rotated pairs closer to their original embeddings. 

\section{Our Approach}
\label{sec:method}

The proposed \ours\ uses a two stage training pipeline; Generation zero (Gen-0) and Generation one (Gen-1). Gen-0 utilizes self-supervision to learn a wider classification manifold, in which the learned embeddings are equivariant to rotation (or another data transformation). Later, during Gen-1, we use Gen-0 model as a teacher and use original (non-rotated) images as anchors to preserve the learned manifold, while rotated version of the images are used to reduce intra-class distances in the embedding space to learn robust and discriminative feature representations.

\subsection{Setting}

Lets assume a neural network \nn\ contains feature embedding parameters \nnP, and classification weights \nnQ. Any input image $\bm{x}$ can be mapped to a feature vector $\bm{v}\in \mathbb{R}^{d}$ by a function \nnFP{}$:\bm{x} \rightarrow \bm{v}$. Consequently, features $\bm{v}$ are mapped to logits $\bm{p}\in \mathbb{R}^{c}$ by another function \nnFQ{}$:\bm{v} \rightarrow \bm{p}$, where $c$ denotes the number of output classes. Hence, conventionally  \nn\ is defined as a composition of these functions, \nn$=$\nnFP{}$\circ$\nnFQ{}. In this work, we introduce another function \nnFR{}, parameterized by \nnR, such that, \nnFR{}$:\bm{p} \rightarrow \bm{q}$, which maps logits $\bm{p}$ to a secondary set of logits $\bm{q}\in \mathbb{R}^{s}$ for self-supervised task (e.g., rotation classification). For each input $\bm{x}$, we automatically obtain labels $r\in \{1,
\ldots, s\}$ for the self-supervision task. Therefore, the complete network can be represented as $F_{\Phi,\Theta,\Psi}=$\nnFR{}$\circ$\nnFQ{}$\circ$\nnFP{}.

We consider a dataset \dset\ with $n$ image-label pairs $\{\bm{x}_i, {y}_i\}_n$ where $y_i \in \{1,\ldots,c\}$. During evaluation, we sample episodes as in classical few-shot learning literature. An episode \deval\ contains, \dsupp\ and \dquery. In an $n$-way $k$-shot setting, \dsupp\ has $k$ number of samples for each of $n$ classes.

\subsection{Generation Zero}

During the first stage (aka Gen-0), a minibatch \mb=\{\x,\y\}\ is randomly sampled from the dataset \dset, which has $m$ number of image-label pairs such that $\text{\x} = \{\bm{x}_i\}_m, \text{\y} = \{{y}_i\}_m$. We first take the images \x\ and apply a transformation function \T\ to create augmented copies of \x. For the sake of brevity, here we consider \T\ as a rotation transformation, however any other suitable transformation can also be considered as we show in our experiments (Sec.~\ref{sec:results_ab}). Applying rotations of $90,180$ and $270$ degrees to \x, we create \xnine, \xoneeight\ and \xtwoseven, respectively. Then we combine all augmented versions of images into a single tensor $\mathbf{\widehat{x}}=\{$\x, \xnine, \xoneeight, \xtwoseven$\}$ whose corresponding class labels are \yhat\ $ \in \mathbb{R}^{4\times m}$. Additionally, one-hot encoded labels \rhat\ $= \{\bm{r}_i \in \mathbb{R}^s\}_{4\times m}$ for the rotation direction are also created, where $s=4$ due to the four rotation directions in our self-supervised task. 

First, we pass \xhat\ through \nnFP{}, resulting in the features \vhat\ $\in \mathbb{R}^{d\times (4\times m)}$. Then, the features are passed through \nnFQ{}\ to get the corresponding logits \phat\ $\in \mathbb{R}^{c\times (4\times m)}$, and finally, the logits are passed through \nnFR{}, to get the rotation logits \qhat\ $\in \mathbb{R}^{s\times (4\times m)}$,
\begin{align*}
    \text{\nnFP{}(\xhat) = \vhat},  \ \ \ \ \ \ \ \
    \text{\nnFQ{}(\vhat) = \phat},  \ \ \ \ \ \ \ \
    \text{\nnFR{}(\phat) = \qhat}.
\end{align*}
We employ, two loss functions to optimize the model in Gen-0: (a) categorical cross entropy loss $\mathcal{L}_{ce}$ between the predicted logits \phat\ and the true labels \yhat, and (b) a self-supervision loss $\mathcal{L}_{ss}$ between the rotation logits \qhat\ and rotation labels \rhat. Note that, self-supervision loss is simply a binary cross entropy loss. These two loss terms are combined with a weighting coefficient $\alpha$ to get our final loss, 
\begin{align*}\small
   \mathcal{L}_{\text{Gen-0}} = \mathcal{L}_{ce} + \alpha \cdot \mathcal{L}_{ss},   \; \text{s.t., }\mathcal{L}_{ce}(\bm{p}, {y}) = -\log\left( \frac{\text{e}^{p_y}}{\sum_j \text{e}^{p_j} } \right), \ 
    \mathcal{L}_{ss}(\bm{q}, r) = -\log\left( \frac{\text{e}^{q_r}}{\sum_j \text{e}^{q_j} } \right).
\end{align*}
The whole process of training the Gen-0 model can be stated as following optimization problem,
\begin{align}
    \min_{\text{\nnP, \nnQ, \nnR}} \;\mathbb{E}_{{\text{\x,\y}}\sim \text{\dset}}\big[\mathcal{L}_{ce}(\text{\nnFPQ{}(\xhat),\ \yhat}) + \alpha \cdot \mathcal{L}_{ss}(\text{\nnFPQR{}(\xhat),\ \rhat})\big].
    \label{eq:main}
\end{align}
The above objective makes sure that the output logits are representative enough to encapsulate information about the input transformation, thereby successfully predicting the amount of rotation applied to the input. This behaviour allows us to maintain diversity in the output space and faithfully estimating the natural data manifold of each object category.

\begin{figure}
    \centering
    \includegraphics[scale=0.14]{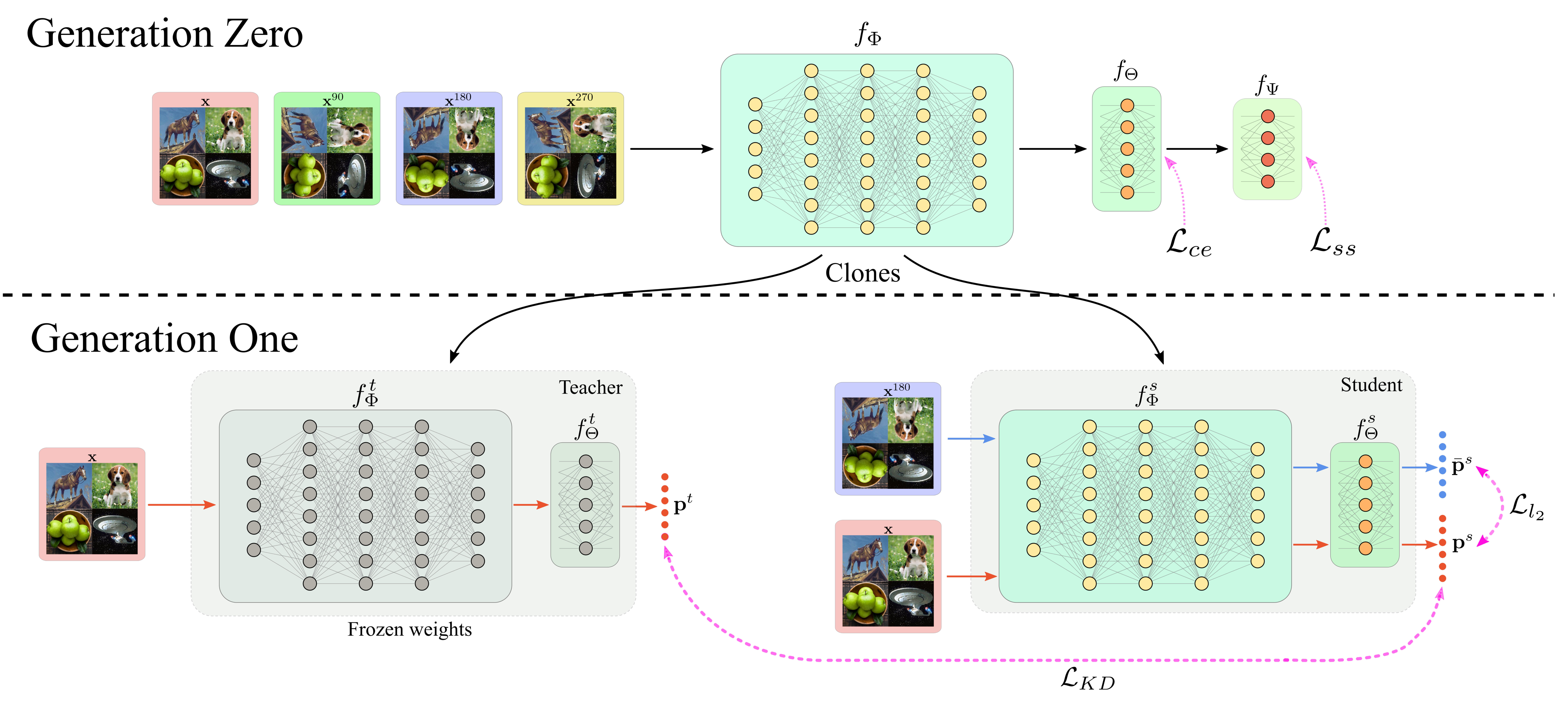}
    \caption{\textit{Overall training process of \ours:} Generation Zero uses multiple rotated versions of the images to train the neural network to predict the class as well as the rotated angle. Then during Generation One, we use original version of the images as anchor points to preserve the manifold while moving the logits for the rotated version closer, to increase the discriminative ability of the network.}
    \label{fig:gen}
\end{figure}

\subsection{Generation One}
Once the Gen-0 model is trained with cross entropy and self-supervision loss functions, we take two clones of the trained model: A teacher model $F^{t}$ and a student model $F^{s}$. Weights of the teacher model are frozen and used only for inference. Again, we sample a minibatch \mb\ from \dset\ and generate a twin \xbar\ $\in$ \xhat$\backslash$\x\ from \x. In this case, a twin \xbar\ is simply a rotated version of \x\ (e.g., \xoneeight). During Gen-1 training, \x\ is used as an anchor point to constrain any changes to the classification manifold. This is enforced by a knowledge distillation~\cite{hinton2015distilling} loss between teacher and student networks. Concurrently, an auxiliary $\ell_2$ loss is employed to bring the embeddings of \x\ and \xbar\ together to enhance feature discriminability while preserving the original output manifold.

Specifically, we first pass \x\ through the teacher network $F^{t} = $ \nnFPQ{t} $\circ$ \nnFR{t} and its logits $\mathbf{p}^t$ are obtained. Then, \x, \xbar\ are passed through the $F^s$ to get their corresponding logits $\mathbf{p}^s$, and  $\bar{\mathbf{p}}^s$ respectively.
\begin{align*}
    \text{\nnFPQ{t}(\x) = $\mathbf{p}^t$},     \ \ \ \ \ \ \ \
    \text{\nnFPQ{s}}(\{\text{\x}, \text{\xbar}\}) = \{\mathbf{p}^s, \bar{\mathbf{p}}^s\}     \ \ \ \ \ \ \ \ \text{s.t., \nnFPQ{}$=$\nnFQ{}$\circ$\nnFP{}.}
\end{align*}
 We use Kullback–Leibler (KL) divergence measure between $\mathbf{p}^t= \{\bm{p}^t_i\}$ and $\mathbf{p}^s = \{\bm{p}^s_i\}$ for knowledge distillation, and apply an $\ell_2$ loss between $\mathbf{p}^s$ and $\bar{\mathbf{p}}^s$ to achieve better discriminability, 
\begin{align*}
    \mathcal{L}_{\text{KD}}(\bm{p}^s, \bm{p}^t, T) = \textrm{KL}\left(\sigma({\bm{p}^s}/{T}), \sigma({\bm{p}^t}/{T})\right), \ \ \ \ \ \ \ \ \ \
    \mathcal{L}_{\ell_2} = \| \bm{p}^s - \bar{\bm{p}}^s\|_2,
\end{align*}
where, $\sigma$ is a softmax function and $T$ is a temperature parameter used to soften the output distribution. Finally, we combine these two loss terms by a coefficient $\beta$ as follows,
\begin{align*}
    \mathcal{L}_{\text{Gen-1}} = \mathcal{L}_{\text{KD}} + \beta \cdot \mathcal{L}_{\ell_2}.
\end{align*}
The overall Gen-1 training process can be stated as the following optimization problem,
\begin{align}
    \min_{\text{\nnP, \nnQ}} \;\mathbb{E}_{{\text{\x,\y}}\sim \text{\dset}}\big[\mathcal{L}_{\text{KD}}(\text{\nnFPQ{s}(\x),\nnFPQ{t}(\x)}) + \beta \cdot \mathcal{L}_{\ell_2}(\text{\nnFPQ{s}(\x),\nnFPQ{s}(\xbar)})\big].
    \label{eq:main}
\end{align}
Note that, in our setting, it is necessary to have the rotation classification head sequentially added to the classification layer, unlike the previous works~\cite{gidaris2019boosting, chen2019self, sun2019unsupervised} which connect rotation classification head directly after the feature embedding layer. This is because, during the Gen-0, we encourage the penultimate layer to encode information about both the image class and its rotation (thus preserving output space diversity), and later in Gen-1, we bring the logits of the rotated pairs closer (to improve discrimination). These benefits are not possible if the rotation head is connected directly to the feature embedding layer 
or if distillation is performed on the feature embeddings.

\begin{algorithm}
\caption{Training procedure of \ours}
\label{alg:train}
\small
\algrenewcommand\algorithmicprocedure{}
\begin{algorithmic}[1]
\Procedure{}{}
\algorithmicrequire{\nnFP{}, \nnFQ{}, \nnFR{}, \dset}
\For {$e$ iterations}  \algorithmiccomment{\emph{Generation Zero training}}
    \While {\text{\mb\ $\sim$ \dset}}
        \State \text{\xnine, \xoneeight, \xtwoseven \ \ $\gets$ \ \ rotate(\x)}
        \State \text{\xhat \ $\gets$ \ \{\x, \xnine, \xoneeight, \xtwoseven\}}, \ \ \text{and} \ \ \text{\yhat \ $\gets$ \ \{\y, \y, \y, \y\}}
        \State \text{\rhat \ $\gets$ \ \{$\mathbf{0}$, $\mathbf{1}$, $\mathbf{2}$, $\mathbf{3}\}$}  
        \algorithmiccomment{where $\mathbf{0}$ is an all zero vector with length $m$}
        \State \text{\vhat  \ $\gets$ \ \nnFP{}(\xhat)}, \ \ \ \ \ \text{\phat  \ $\gets$ \ \nnFQ{}(\vhat)}, \ \ \ \ \ \text{\qhat  \ $\gets$  \ \nnFR{}(\phat)}
        \State \text{$\mathcal{L}_{0}$  \ \ $\gets$ \ \  $\mathcal{L}_{ce}$(\phat, \yhat) + $\alpha$ $\cdot$ $\mathcal{L}_{ss}$(\qhat, \rhat)}
        \State \text{$\{$ \nnP, \nnQ, \nnR  $\}$ \ \ $\gets$ \ \ $\{$ \nnP, \nnQ, \nnR  $\}$} - \text{$\nabla_{\text{{$\{$ \nnP, \nnQ, \nnR  $\}$}}} \mathcal{L}_0$ }
    \EndWhile
\EndFor
\State ${F}^t, {F}^s \gets {F}$
\For {$e$ iterations}  \algorithmiccomment{\emph{Generation One training}}
    \While {\text{\mb\ $\sim$ \dset}}
        \State \text{\xoneeight \ \ $\gets$ \ \ rotate(\x)}
        \State \text{$\mathbf{p}^t$ \ $\gets$ \ \nnFPQ{t}(\x)}, \ \ \ \text{$\{\mathbf{p}^s, \bar{\mathbf{p}}^s\}$ \ $\gets$ \ \nnFPQ{s}(\{\x, \xoneeight\})}
        \State \text{$\mathcal{L}_{1}$  \ \ $\gets$ \ \  $\mathcal{L}_{\text{KD}}$($\mathbf{p}^s$, $\mathbf{p}^t$) + $\beta$ $\cdot$ $\mathcal{L}_{\ell_2}$($\mathbf{p}^s$, $\bar{\mathbf{p}}^s$)}
        \State \text{$\{$ \nnP, \nnQ$\}$ \ \ $\gets$ \ \ $\{$ \nnP, \nnQ $\} - \nabla_{\text{{$\{$ \nnP, \nnQ$\}$}}} \mathcal{L}_1$ }
    \EndWhile
\EndFor
\Return \text{\nnFPQ{s}}
\EndProcedure
\end{algorithmic}
\end{algorithm}

\subsection{Evaluation}

During evaluation, a held out part of the dataset is used to sample tasks. This comprises of a support set and a query set $\{\text{\dsupp, \dquery}\}$. \dsupp\ has image-label pairs $\{\mathbf{x}_{supp}, \mathbf{y}_{supp}\}$, while \dquery\ comprise of an image tensor $\mathbf{x}_{query}$. Both $\mathbf{x}_{supp}$ and $\mathbf{x}_{query}$ are fed to the final trained \nnFPs\ model to get the feature embeddings $\mathbf{v}_{supp}$ and $\mathbf{v}_{query}$, respectively. We use a simple logistic regression classifier~\cite{tian2020rethink,bertinetto2018meta} to map the labels from support set to query set. The embeddings are normalized onto a unit sphere~\cite{tian2020rethink}. We randomly sample 600 tasks, and report mean classification accuracy with 95\% confidence interval. Unlike popular meta-learning algorithms (e.g., \cite{finn2017model,li2017meta}), we do not need to train multiple models for different values of $n$ and $k$ in $n$-way, $k$-shot classification. Since, the classification is disentangled from feature learning in our case, the same model can be used to evaluate for any value of $n$ and $k$ in FSL.

\section{Experiments and Results}
\label{sec:results}

We comprehensively compare our method on four benchmark few-shot learning datasets i.e., miniImageNet~\cite{vinyals2016matching}, tieredImageNet~\cite{ren2018metalearning}, CIFAR-FS~\cite{bertinetto2018meta} and FC100~\cite{oreshkin2018tadam}. Additionally, we provide an extensive ablation study to investigate the individual contributions of different components (Sec.~\ref{sec:results_ab}).

\textbf{Implementation Details:}
To be consistent with previous methods~\cite{tian2020rethink, mishra2018a, oreshkin2018tadam, lee2019meta}, we use ResNet-12 as the backbone in our experiments. The backbone architecture contains 4 residual blocks of $64$, $160$, $320$, $640$ filters as in ~\cite{tian2020rethink, ravichandran2019few, lee2019meta}, each with $3 \times 3$ convolutions. A $2 \times 2$ max pooling operation is applied after the first 3 blocks and a global average pooling is applied after the last block. Additionally, a $4$ neuron fully-connected layer is added after the final classification layer.

We use SGD with an initial learning rate of $0.05$, momentum of $0.9$, and a weight decay of $5e{-4}$. The learning rate is reduced after epoch 60 by a factor of $0.1$. Gen-0 and Gen-1 models on CIFAR-FS are trained for 65 epochs, while rest of the models are trained for 8 epochs only. consistent with previous approaches~\cite{finn2017model, tian2020rethink, rusu2018metalearning}, random crop, color jittering and random horizontal flip are applied for data augmentation during training.  Further, the hyper-parameters $\alpha, \beta$ are tuned on a validation set, and we used the same value of $4.0$ as in~\cite{tian2020rethink} for temperature coefficient $T$ during distillation.

\textbf{Datasets: }We evaluate our method on four widely used FSL benchmarks. These include two datasets which are subsets of the ImageNet i.e., miniImageNet~\cite{vinyals2016matching} and tieredImageNet~\cite{ren2018metalearning}, and the other two which are splits of CIFAR100 i.e., CIFAR-FS~\cite{bertinetto2018meta} and FC100~\cite{oreshkin2018tadam}. For miniImageNet~\cite{vinyals2016matching}, we use the split proposed in~\cite{ravi2016optimization}, with 64, 16 and 20 classes for training, validation and testing. The tieredImageNet~\cite{ren2018metalearning} contains 608 classes which are semantically grouped into 34 high-level classes, that are further divided into 20, 6 and 8 for training, validation, and test splits, thus making the splits more diverse. CIFAR-FS~\cite{bertinetto2018meta} contains a random split of 100 classes into 64, 16 and 20 for training, validation, and testing, while FC100~\cite{oreshkin2018tadam} uses a split similar to tieredImageNet, making the splits more diverse. FC100 has 60, 20, 20 classes for training, validation, and testing respectively.

\begin{table*}[htp]
\setlength{\tabcolsep}{0.3cm}
    \centering
        \begin{center}
        \scalebox{0.8}{
        \begin{tabular}{c c c c c c}
        \toprule[0.4mm]
 \rowcolor{mygray}
 & & \multicolumn{2}{c}{miniImageNet, 5-way} & \multicolumn{2}{c}{tieredImageNet, 5-way} \\
\rowcolor{mygray} \multirow{-2}{*}{Method} & \multirow{-2}{*}{Backbone} & 1-shot & 5-shot &  1-shot & 5-shot  \\ \midrule
MAML~\cite{finn2017model}                      & 32-32-32-32   & 48.70 $\pm$ 1.84    & 63.11 $\pm$ 0.92    &  51.67 $\pm$ 1.81  &  70.30 $\pm$ 1.75 \\
Matching Networks~\cite{vinyals2016matching}   & 64-64-64-64   & 43.56 $\pm$ 0.84    & 55.31 $\pm$ 0.73    & \ \ - & \ \ - \\
IMP~\cite{pmlr-v97-allen19b}                   & 64-64-64-64   & 49.2 \ \ $\pm$ 0.7 \ \    & 64.7 \ \ $\pm$ 0.7 \ \ & \ \ - & \ \ - \\
Prototypical Networks$^\dagger$~\cite{snell2017prototypical}   & 64-64-64-64  & 49.42 $\pm$ 0.78 & 68.20 $\pm$ 0.66 & 53.31 $\pm$ 0.89 & 72.69 $\pm$ 0.74 \\
TAML~\cite{jamal2019task}                      & 64-64-64-64   & 51.77 $\pm$ 1.86    & 66.05 $\pm$ 0.85    & \ \ - & \ \ - \\
SAML~\cite{hao2019collect}                     & 64-64-64-64   & 52.22 $\pm$ n/a \ \ & 66.49 $\pm$ n/a \ \ & \ \ - & \ \ - \\
GCR~\cite{li2019few}                           & 64-64-64-64   & 53.21 $\pm$ 0.80    & 72.34 $\pm$ 0.64    & \ \ - & \ \ - \\
KTN(Visual)~\cite{peng2019few}                 & 64-64-64-64   & 54.61 $\pm$ 0.80    & 71.21 $\pm$ 0.66    & \ \ - & \ \ - \\
PARN~\cite{wu2019parn}                         & 64-64-64-64   & 55.22 $\pm$ 0.84    & 71.55 $\pm$ 0.66    & \ \ - & \ \ - \\
Dynamic Few-shot~\cite{gidaris2018dynamic}     & 64-64-128-128 & 56.20 $\pm$ 0.86    & 73.00 $\pm$ 0.64    & \ \ - & \ \ - \\
Relation Networks~\cite{sung2018learning}      & 64-96-128-256 & 50.44 $\pm$ 0.82    & 65.32 $\pm$ 0.70    & 54.48 $\pm$ 0.93 & 71.32 $\pm$ 0.78 \\
R2D2~\cite{bertinetto2018meta}                 & 96-192-384-512& 51.2 \ \ $\pm$ 0.6 \ \ & 68.8 \ \ $\pm$ 0.1 \ \ & \ \ - & \ \ - \\
SNAIL~\cite{mishra2018a}                       & ResNet-12     & 55.71 $\pm$ 0.99    & 68.88 $\pm$ 0.92    & \ \ - & \ \ - \\
AdaResNet~\cite{pmlr-v80-munkhdalai18a}        & ResNet-12     & 56.88 $\pm$ 0.62    & 71.94 $\pm$ 0.57    & \ \ - & \ \ - \\
TADAM~\cite{oreshkin2018tadam}                 & ResNet-12     & 58.50 $\pm$ 0.30    & 76.70 $\pm$ 0.30    & \ \ - & \ \ - \\
Shot-Free~\cite{ravichandran2019few}           & ResNet-12     & 59.04 $\pm$ n/a \ \ & 77.64 $\pm$ n/a \ \ & 63.52 $\pm$ n/a \ \ & 82.59 $\pm$ n/a \ \ \\
TEWAM~\cite{qiao2019transductive}              & ResNet-12     & 60.07 $\pm$ n/a \ \ & 75.90 $\pm$ n/a \ \ & \ \ - & \ \ - \\
MTL~\cite{sun2019meta}                         & ResNet-12     & 61.20 $\pm$ 1.80    & 75.50 $\pm$ 0.80    & \ \ - & \ \ - \\
Variational FSL~\cite{schonfeld2019generalized}& ResNet-12     & 61.23 $\pm$ 0.26    & 77.69 $\pm$ 0.17    & \ \ - & \ \ - \\
MetaOptNet~\cite{lee2019meta}                  & ResNet-12     & 62.64 $\pm$ 0.61    & 78.63 $\pm$ 0.46    & 65.99 $\pm$ 0.72 & 81.56 $\pm$ 0.53 \\
Diversity w/ Cooperation~\cite{dvornik2019diversity}& ResNet-18& 59.48 $\pm$ 0.65    & 75.62 $\pm$ 0.48    & \ \ - & \ \ - \\
Boosting~\cite{gidaris2019boosting}            & WRN-28-10     & 63.77 $\pm$ 0.45    & 80.70 $\pm$ 0.33    & 70.53 $\pm$ 0.51 & 84.98 $\pm$ 0.36  \\
Fine-tuning~\cite{Dhillon2020A}                & WRN-28-10     & 57.73 $\pm$ 0.62    & 78.17 $\pm$ 0.49    & 66.58 $\pm$ 0.70 & 85.55 $\pm$ 0.48 \\
LEO-trainval$^\dagger$~\cite{rusu2018metalearning} & WRN-28-10 & 61.76 $\pm$ 0.08    & 77.59 $\pm$ 0.12    & 66.33 $\pm$ 0.05 & 81.44 $\pm$ 0.09 \\
RFS-simple~\cite{tian2020rethink}              & ResNet-12     & 62.02 $\pm$ 0.63    & 79.64 $\pm$ 0.44    & 69.74 $\pm$ 0.72 & 84.41 $\pm$ 0.55 \\
RFS-distill~\cite{tian2020rethink}             & ResNet-12     & 64.82 $\pm$ 0.60    & 82.14 $\pm$ 0.43    & 71.52 $\pm$ 0.69 & 86.03 $\pm$ 0.49 \\ \midrule
\ours-GEN0                                   & ResNet-12     & 65.93 $\pm$ 0.81    & 83.15 $\pm$ 0.54    & 71.69 $\pm$ 0.91 & \textbf{86.66 $\pm$ 0.60} \\ 
\ours-GEN1                                     & ResNet-12     & \textbf{67.04 $\pm$ 0.85}    & \textbf{83.54 $\pm$ 0.54}    & \textbf{72.03 $\pm$ 0.91} & {86.50 $\pm$ 0.58} \\
\bottomrule[0.4mm]
\end{tabular}}
\end{center}\vspace{-0.5em}
\caption{FSL results on miniImageNet~\cite{vinyals2016matching} and tieredImageNet~\cite{ren2018metalearning} datasets, with mean accurcy and 95\% confidence interval. $^\dagger$results obtained by training on train+val sets. Table is an extended version from~\cite{tian2020rethink}. }
\label{tbl:results1}
\end{table*}

\subsection{Few-shot learning results}
Our results shown in Table~\ref{tbl:results1} and~\ref{tbl:results2} suggest that the proposed method consistently outperforms the current methods on all four datasets. Even, our Gen-0 alone performs better than the current state-of-the-art (SOTA) methods by a considerable margin. For example, \ours\ Gen-0 model surpasses SOTA performance on miniImageNet by $\sim$1\% on both 5-way 1-shot and 5-way 5-shot tasks. The same can be observed on other datasets. Compared to RFS-simple~\cite{tian2020rethink} (similar to our Gen-0), \ours shows an improvement of $3.91\%$ on 5-way 1-shot and $3.51\%$ on 5-way 5-shot learning. The same trend can be observed across other evaluated datasets with consistent $2$-$3\%$ gains over RFS-simple. 
This is due to the novel self-supervision which enables SKD to learn diverse and generalizable embedding space. 

Gen-1 incorporates knowledge distillation and proves even more effective compared with Gen-0. On miniImageNet, we achieve $67.04\%$ and $83.54\%$ on 5-way 1-shot and 5-way 5-shot learning tasks, respectively. These are gains of $2.22\%$ and $1.4\%$ on 5-way 1-shot and 5-way 5-shot tasks. Similar consistent gains of $2$-$3\%$ over SOTA results can be observed across other evaluated datasets. Note that, RFS-distill~\cite{tian2020rethink} uses multiple iterations (up to 3-4 generations) for model distillation, while \ours only uses a single generation for the distillation. We attribute our gain to the way we use knowledge distillation to constrain changes in the embedding space, while minimizing the embedding distance between images and their rotated pairs, thus enhancing representation capabilities of the model.

\begin{table*}[!t]
\setlength{\tabcolsep}{0.4cm}
    \centering
        \begin{center}
        \scalebox{0.8}{
        \begin{tabular}{c c c c c c}
        \toprule[0.4mm]
\rowcolor{mygray} &  & \multicolumn{2}{c}{CIFAR-FS, 5-way} & \multicolumn{2}{c}{FC100, 5-way} \\
\rowcolor{mygray} \multirow{-2}{*}{Method} & \multirow{-2}{*}{Backbone} & 1-shot & 5-shot &  1-shot & 5-shot  \\ \midrule
MAML~\cite{finn2017model}                      & 32-32-32-32   & 58.9 $\pm$ 1.9    & 71.5 $\pm$ 1.0 &  \ \ - & \ \ - \\
Prototypical Networks$^\dagger$~\cite{snell2017prototypical}   & 64-64-64-64 & 55.5 $\pm$ 0.7  & 72.0 $\pm$ 0.6 & 35.3 $\pm$ 0.6 & 48.6 $\pm$ 0.6 \\
Relation Networks~\cite{sung2018learning}      & 64-96-128-256 & 55.0 $\pm$ 1.0    & 69.3 $\pm$ 0.8 & \ \ - & \ \ - \\
R2D2~\cite{bertinetto2018meta}                 & 96-192-384-512& 65.3 $\pm$ 0.2    & 79.4 $\pm$ 0.1 & \ \ - & \ \ - \\
TADAM~\cite{oreshkin2018tadam}                 & ResNet-12     & \ \ -             & \ \ -          & 40.1 $\pm$ 0.4 & 56.1 $\pm$ 0.4 \\
Shot-Free~\cite{ravichandran2019few}           & ResNet-12     & 69.2 $\pm$ n/a    & 84.7 $\pm$ n/a & \ \ - & \ \ - \\
TEWAM~\cite{qiao2019transductive}              & ResNet-12     & 70.4 $\pm$ n/a    & 81.3 $\pm$ n/a & \ \ - & \ \ - \\
Prototypical Networks$^\dagger$~\cite{snell2017prototypical}   & ResNet-12  & 72.2 $\pm$ 0.7 & 83.5 $\pm$ 0.5 & 37.5 $\pm$ 0.6 & 52.5 $\pm$ 0.6 \\
Boosting~\cite{gidaris2019boosting}            & WRN-28-10     & 73.6 $\pm$ 0.3    & 86.0 $\pm$ 0.2 & \ \ - & \ \ - \\
MetaOptNet~\cite{lee2019meta}                  & ResNet-12     & 72.6 $\pm$ 0.7    & 84.3 $\pm$ 0.5 & 41.1 $\pm$ 0.6 & 55.5 $\pm$ 0.6 \\
RFS-simple~\cite{tian2020rethink}              & ResNet-12     & 71.5 $\pm$ 0.8    & 86.0 $\pm$ 0.5 & 42.6 $\pm$ 0.7 & 59.1 $\pm$ 0.6 \\
RFS-distill~\cite{tian2020rethink}             & ResNet-12     & 73.9 $\pm$ 0.8    & 86.9 $\pm$ 0.5 & 44.6 $\pm$ 0.7 & 60.9 $\pm$ 0.6 \\ \midrule
\ours-GEN0                                     & ResNet-12     & 74.5 $\pm$ 0.9    & 88.0 $\pm$ 0.6 & 45.3 $\pm$ 0.8 & 62.2 $\pm$ 0.7 \\
\ours-GEN1                                     & ResNet-12     & \textbf{76.9 $\pm$ 0.9}    & \textbf{88.9 $\pm$ 0.6} & \textbf{46.5 $\pm$ 0.8} & \textbf{63.1 $\pm$ 0.7} \\
\bottomrule[0.4mm]
\end{tabular}}
\end{center}\vspace{-0.5em}
\caption{FSL on CIFAR-FS~\cite{bertinetto2018meta} and FC100~\cite{oreshkin2018tadam} datasets, with mean accurcy and 95\% confidence interval. $^\dagger$results obtained by training on train+val sets. Table is an extended version from~\cite{tian2020rethink}. }
\label{tbl:results2}
\end{table*}

\subsection{Ablation Studies and Analysis}
\label{sec:results_ab}

\noindent\textbf{Choices of loss function:}  We study the impact of different contributions by progressively integrating them into our proposed method. To this end, we first evaluate our method with and without the self-supervision loss. If we train the Gen-0 with only cross entropy loss, which is same as RFS-simple~\cite{tian2020rethink}, the model achieve $71.5 \pm 0.8\%$ and $62.02 \pm 0.63\%$ on 5-way 1-shot task on CIFAR-FS and miniImageNet, respectively. Then, if we train the Gen-0 with additional self supervision, the model performance improves to $74.5 \pm 0.9\%$ and $65.93 \pm 0.81\%$. This shows an absolute gain of $3.0\%$ and $3.91\%$, by incorporating our proposed self-supervision. Additionally, if we only keep knowledge distillation for Gen-1, we can see that  self-supervision for Gen-0 has a clear impact on next generation. As shown in Table~\ref{tbl:ab_loss}, self-supervision at Gen-0 is responsible for $2\%$ performance improvement on Gen-1. Further, during Gen-1, the advantage of using the $\mathcal{L}_{\ell_2}$ loss to bring logits of rotated augmentations closer, is demonstrated in Table~\ref{tbl:ab_loss}. We can see that, even for both Gen-0 models trained on $\mathcal{L}_{ce}$ and $\mathcal{L}_{ce} + \alpha \mathcal{L}_{ss}$, addition of $\mathcal{L}_{\ell_2}$ loss during Gen-1 gives about $\sim1\%$ gain compared with using knowledge distillation only. These emprical evaluations clearly establish individual importance of different contributions (self-supervision, knowledge distillation and ensuring proximity of augmented versions of the image in output space) in our proposed two stage approach.

\noindent\textbf{Choices of self supervision:} We further investigate different choices of self-supervision. Instead of rotations based self-supervision, we use a $2 \times 2$ crop of an image, and train the final classifier to predict the correct crop quadrant~\cite{sun2019unsupervised}. The results in Table~\ref{tbl:ab_ss} show that the crop-based self-supervision method can also surpass the SOTA FSL methods, though it performs slightly lower than the rotations based self-supervision. We further experiment with simCLR loss \cite{chen2020simple}, which also aims to bring augmented pairs closer together, along-with knowledge distillation during Gen-1. Our experiments show that simCLR only achieves $75.0 \pm 0.8$ and $88.2 \pm 0.6\%$ on 1 and 5 shot tasks respectively.

\begin{figure*}[!t]
    \centering
    \begin{subfigure}{0.4\textwidth}
    \label{fig:meta_agnostic}
        \centering
        \includegraphics[width=\textwidth]{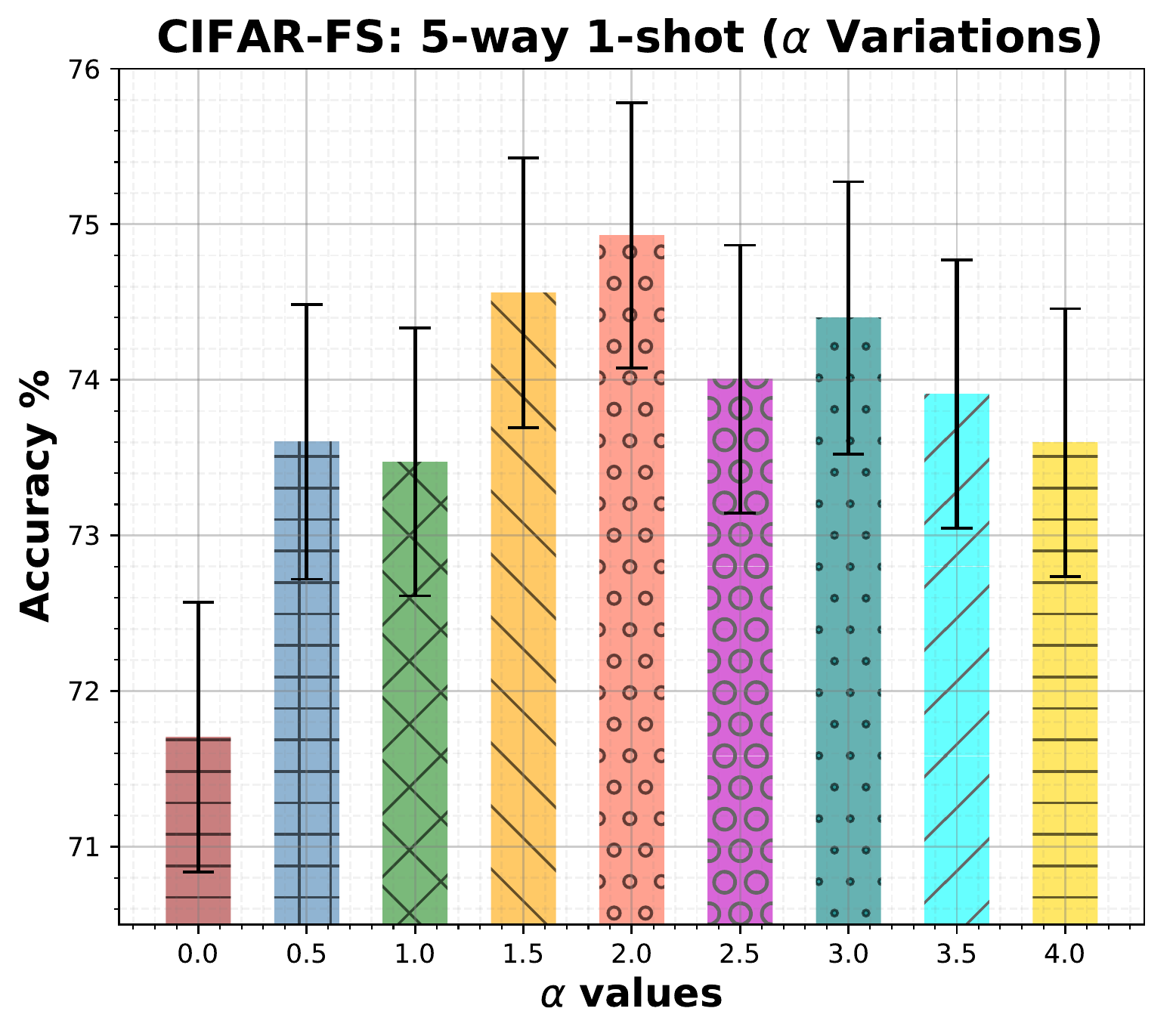}
    \end{subfigure}%
    \;\;\;
    \begin{subfigure}{.4\textwidth}
        \centering
        \includegraphics[width=\textwidth]{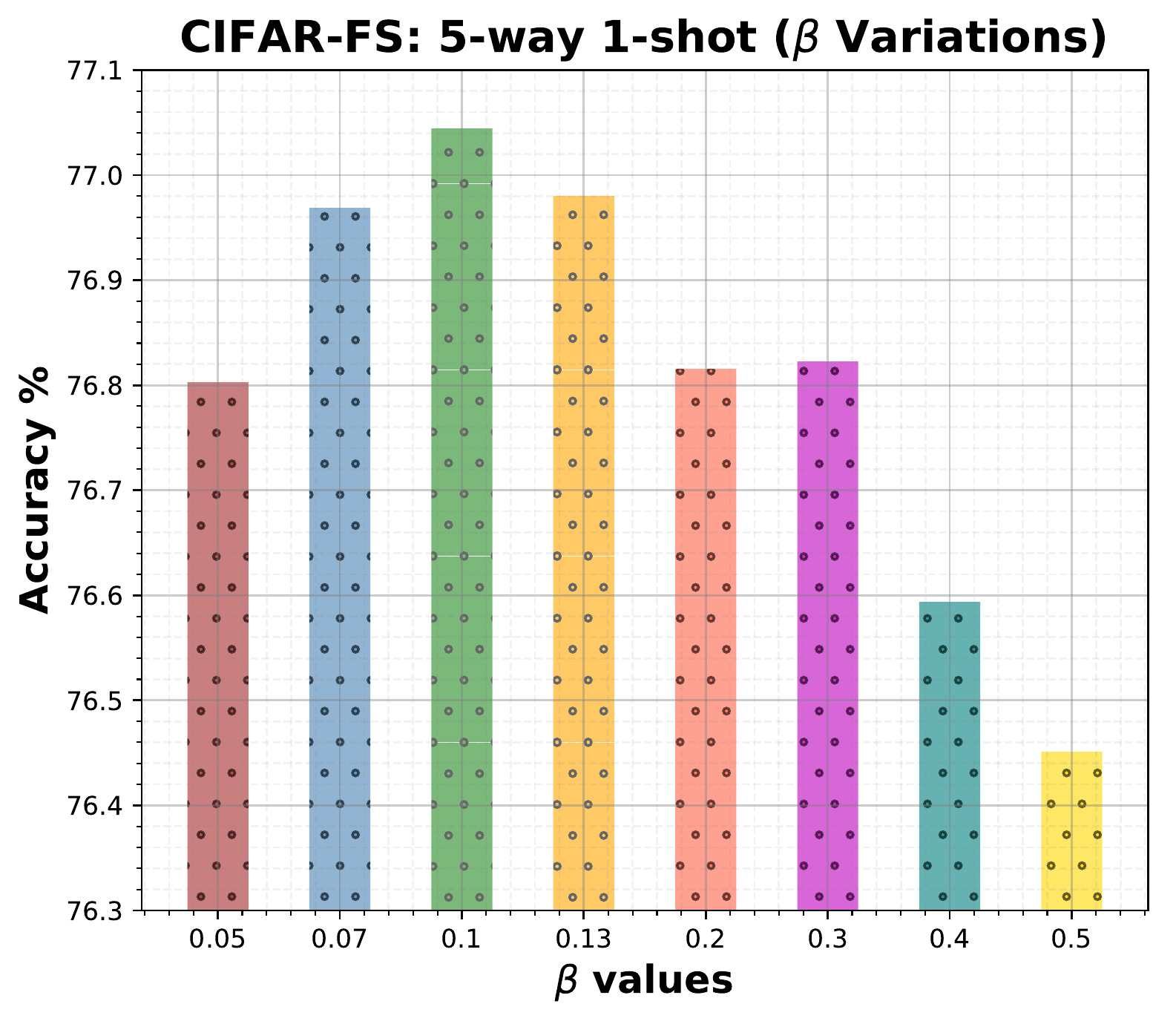}
    \end{subfigure}%
    \vspace{-0.2cm}
    \caption{Ablation study on the sensitivity of the loss coefficient hyper-parameters $\alpha$ and $\beta$. } 
    \label{fig:ab_alphabeta}
\end{figure*}

\noindent\textbf{Variations of $\alpha$:} During Gen-0 of \ours, $\alpha$ controls the contribution of self-supervision over classification. Fig.~\ref{fig:ab_alphabeta} (\emph{left}) shows the Gen-0 performance by changing $\alpha$. We observe that the performance increases from 0 till 2, and then decreases. The results indicate that the performance is not sensitive to the values of  $\alpha$. It is important to note that Gen-0 without self-supervision i.e. $\alpha=0$ performs the lowest compared with other values of $\alpha$, thus establishing the importance of self-supervision.

\noindent\textbf{Variations of $\beta$:} At Gen-1, we again use a coefficient $\beta$ to control the contribution of loss $\mathcal{L}_{\ell_2}$ over knowledge distillation. From results in Fig.~\ref{fig:ab_alphabeta} (\emph{right}), we observe a similar trend as for the case of $\alpha$, that the performance first improves for $0\leq\beta\leq0.1$, and then decreases with larger values of $\beta$. However, even if we change $\beta$ from $0.1$ to $0.5$, the performance drops only by $\sim0.6\%$. Note that, the performance on CIFAR-FS, on 5-way 1-shot without $\mathcal{L}_{\ell_2}$ loss is only $75.6 \pm 0.9\%$, which is the lowest compared with other values of $\beta$, showing the importance of $\mathcal{L}_{\ell_2}$.

\noindent\textbf{Time Complexity}
Lets assume $T$ is the time required for training one generation. RFS~\cite{tian2020rethink} has the time complexity of $\mathcal{O}(n\times T)$, where $n$ is the number of generations, which is usually about 3-4. However, our complexity is $\mathcal{O}(2 \times T)$. Note that, additional rotation augmentations do not affect the training time $T$ much, with parallel computing on GPUs. Also, we generally train Gen-1 model for less number of epochs than Gen-0.
Using a single Tesla V100 GPU on CIFAR-FS,  for the first generation, both RFS and \ours\ take approximately the same time, i.e., $T=88$ minutes.  The complete training time on CIFAR-FS of RFS is $\sim 4$ hours, while \ours only takes $\sim 2$ hours.

\begin{table*}
    \centering \setlength{\tabcolsep}{0.3cm}
        \begin{center}
        \scalebox{0.8}{
        \begin{tabular}{c l c c c c c}
        \toprule[0.4mm]
\rowcolor{mygray}  &  & \multicolumn{2}{c}{CIFAR-FS, 5-way} & \multicolumn{2}{c}{miniImageNet, 5-way} \\
\rowcolor{mygray} \multirow{-2}{*}{Generation} & \multirow{-2}{*}{Loss Function} & 1-shot & 5-shot &  1-shot & 5-shot  \\ \midrule
\multirow{2}{*}{GEN-0} & $\mathcal{L}_{CE}$                                                     & 71.5 $\pm$ 0.8    & 86.0 $\pm$ 0.5 &  62.02 $\pm$ 0.63    & 79.64 $\pm$ 0.44 \\
                       & $\mathcal{L}_{CE} + \alpha \mathcal{L}_{SS}$                           & 74.5 $\pm$ 0.9    & 88.0 $\pm$ 0.6 &  65.93 $\pm$ 0.81    & 83.15 $\pm$ 0.54 \\ \midrule
\multirow{4}{*}{GEN-1} & $\mathcal{L}_{CE} \to \mathcal{L}_{KD}$                                & 73.9 $\pm$ 0.8    & 86.9 $\pm$ 0.5 &  64.82 $\pm$ 0.60    & 82.14 $\pm$ 0.43 \\
                       & $\mathcal{L}_{CE} \to \mathcal{L}_{KD} + \beta \mathcal{L}_{\ell_2}$      & 74.9 $\pm$ 1.0    & 87.6 $\pm$ 0.6 &   64.76 $\pm$ 0.84    & 81.84 $\pm$ 0.54 \\
                       & $\mathcal{L}_{CE} + \alpha \mathcal{L}_{SS} \to \mathcal{L}_{KD}$                                &  75.6 $\pm$ 0.9    & 88.7 $\pm$ 0.6 &  66.48 $\pm$ 0.84    & \textbf{83.64 $\pm$ 0.53} \\
                       & $\mathcal{L}_{CE} + \alpha \mathcal{L}_{SS} \to \mathcal{L}_{KD} + \beta \mathcal{L}_{\ell_2}$      &  \textbf{76.9 $\pm$ 0.9}    & \textbf{88.9 $\pm$ 0.6} &  \textbf{67.04 $\pm$ 0.85}    & 83.54 $\pm$ 0.54 \\                       
\bottomrule[0.4mm]
\end{tabular} }
\end{center}\vspace{-0.5em}
\caption{FSL results on CIFAR-FS~\cite{bertinetto2018meta} and FC100~\cite{oreshkin2018tadam}, with different combinations of loss functions for Gen-0 and Gen-1. For Gen-1, the loss functions on the left side of the arrow were used to train the Gen-0 model. }
\label{tbl:ab_loss}
\end{table*}

\begin{table*}
    \centering
\begin{minipage}{0.34\textwidth}
\vspace{0.2cm}
\caption{Few-shot learning results on CIFAR-FS~\cite{bertinetto2018meta} and FC100~\cite{oreshkin2018tadam} datasets, with a comparison to no self-supervision vs finding the rotation and finding the location of a patch as self-supervision method.  }
\label{tbl:ab_ss}
\end{minipage}
\hfill
\begin{minipage}{0.65\textwidth}
\vspace{-0.3cm}
\setlength{\tabcolsep}{0.2cm}
        \scalebox{0.8}{
        \begin{tabular}{c l c c c c c}
        \toprule[0.4mm]
\rowcolor{mygray} & \multicolumn{2}{c}{Generation 0, 5-way} & \multicolumn{2}{c}{Generation 1, 5-way} \\
\rowcolor{mygray} \multirow{-2}{*}{Self-supervision Type}  & 1-shot & 5-shot &  1-shot & 5-shot  \\ \midrule
None     & 71.5 $\pm$ 0.8    & 86.0 $\pm$ 0.5    & 73.9 $\pm$ 0.8    & 86.9 $\pm$ 0.5  \\  \midrule
Rotation & \textbf{74.5 $\pm$ 0.9}    & \textbf{88.0 $\pm$ 0.6}    & \textbf{76.9 $\pm$ 0.9}    & \textbf{88.9 $\pm$ 0.6}  \\ 
Location & 74.1 $\pm$ 0.9    & \textbf{88.0 $\pm$ 0.6}    & 76.2 $\pm$ 0.9    & 87.8 $\pm$ 0.6  \\ 
\bottomrule[0.4mm]
        \end{tabular}
        }
\end{minipage}
\vspace{-1.5em}
\end{table*}

\section{Conclusion}
Deep learning models can easily overfit on the scarce data available in FSL settings. To enhance generalizability, existing approaches regularize the model to preserve margins or encode high-level learning behaviour via meta-learning. In this work, we take a different approach and propose to learn the true output classification manifold via self-supervised learning. Our approach operates in two phases: first, the model learns to classify inputs such that the diversity in the outputs is not lost, thereby avoiding overfitting and modeling the natural output manifold structure. Once this structure is learned, our approach trains a student model that preserves the original output manifold structure while jointly maximizing the discriminability of learned representations. Our results on four popular benchmarks show the benefit of our approach where it establishes a new state-of-the-art for FSL.

\section*{Broader Impact}
This research aims to equip machines with capabilities to learn new concepts using only a few examples. Developing machine learning models which can generalize to a large number of object classes using only a few examples has numerous potential applications with a positive impact on society. Examples include enabling visually impaired individuals to understand the environment around them and enhancing the capabilities of robots being used in healthcare and elderly care facilities. It has the potential to reduce expensive and laborious data acquisition and annotation effort required to learn models in domains including image classification, retrieval, language modelling and object detection. However, we must be cautious that the few shot learning techniques can be misused by authoritarian government agencies which can compromise an individual's privacy. 

{\small
\bibliographystyle{abbrv}
\bibliography{ref}}

\end{document}